# Voice-assisted Image Labelling for Endoscopic Ultrasound Classification using Neural Networks

Ester Bonmati, Yipeng Hu, Alexander Grimwood, Gavin J. Johnson, George Goodchild, Margaret G. Keane, Kurinchi Gurusamy, Brian Davidson, Matthew J. Clarkson, Stephen P. Pereira, Dean C. Barratt

*Abstract*—Ultrasound imaging is a commonly used technology for visualising patient anatomy in real-time during diagnostic and therapeutic procedures. High operator dependency and low reproducibility make ultrasound imaging and interpretation challenging with a steep learning curve. Automatic image classification using deep learning has the potential to overcome some of these challenges by supporting ultrasound training in novices, as well as aiding ultrasound image interpretation in patient with complex pathology for more experienced practitioners. However, the use of deep learning methods requires a large amount of data in order to provide accurate results. Labelling large ultrasound datasets is a challenging task because labels are retrospectively assigned to 2D images without the 3D spatial context available in vivo or that would be inferred while visually tracking structures between frames during the procedure. In this work, we propose a multi-modal convolutional neural network (CNN) architecture that labels endoscopic ultrasound (EUS) images from raw verbal comments provided by a clinician during the procedure. We use a CNN composed of two branches, one for voice data and another for image data, which are joined to predict image labels from the spoken names of anatomical landmarks. The network was trained using recorded verbal comments from expert operators. Our results show a prediction accuracy of 76% at image level on a dataset with 5 different labels. We conclude that the addition of spoken commentaries can increase the performance of ultrasound image classification, and eliminate the burden of manually labelling large EUS datasets necessary for deep learning applications.

*Index Terms*— Automatic labelling, classification, deep learning, ultrasound, voice.

## I. INTRODUCTION

ULTRASOUND (US) imaging is a safe, non-invasive and cost-effective technology for visualising patient anatomy in real-time. However, US scanning is highly operator-dependent and images can be difficult to interpret, requiring extensive training with a long learning curve [1]. To address these challenges in US-guided procedures, several simulators have been proposed. However, even after completing the recommended training, a clinician may find it difficult to perform the examination confidently [1]. In addition, outside of direct supervision by another clinician, ways of guiding inexperienced clinicians are scarce [2]. Therefore, with the ubiquitous use of US imaging, there is a need to develop tools that can assist clinicians during these procedures.

The use of automatic image classification in real-time during an US procedure could potentially increase the confidence of a clinician by labelling images of anatomical landmarks of interest and, in turn, decrease the procedure time. In the literature, deep learning models have shown successful results in several visual recognition tasks on US images such as: classification [3]–[5], segmentation [6]–[9], object detection [10], [11] and plane detection [12]. However, a challenge in using deep learning-based methods for US is the acquisition of a sufficient number of labelled intraoperative data for training. In this regard, several approaches have been proposed when annotations are scarce or weak [13]. The acquisition of labelled data is especially difficult for procedures such as endoscopic ultrasound (EUS) because labelling anatomical landmarks in real-time is time-consuming and can be disruptive to the examination. Image labels annotated retrospectively, on the other hand, may be inaccurate as the 3D spatial and temporal context required to confidently localise the 2D images is often lost.

Studies on speech recognition and natural language processing have shown that convolutional neural networks (CNNs) [14] and recurrent neural networks (RNNs) [15] can successfully transcribe voice to text. The use of text has been explored for US image captioning, either using medical reports [16], [17] or voice commentaries on a retrospective dataset [18]. In order to be less text-dependant, in the computer vision community, there is an interest in machine learning methods that learn directly from voice signals, rather than transcripts [19]–[23]. For example, Harwath et al.[20], proposed a novel multi-modal method for image retrieval that uses spoken captions on real images to assign a similarity score to each image, demonstrating the possibility to learn semantic correspondences from audio and image pairings. This approach is especially clinically relevant because trainees learning EUS are routinely asked to describe multiple anatomical landmarks in US images during scans, while experienced clinicians also

This work was supported by Cancer Research UK (CRUK) Multidisciplinary Award (C28070/A19985) and by the Wellcome/EPSRC Centre for Interventional and Surgical Sciences (WEISS) (203145/Z/16/Z).

E. Bonmati, Y. Hu, A. Grimwood, M. J. Clarkson and D. C. Barratt are with the Wellcome / EPSRC Centre for Interventional and Surgical Sciences, University College London, London, UK, and the UCL Centre for Medical Image Computing, University College London, London, UK (e-mail: e.bonmati@ucl.ac.uk).
G. J. Johnson, G. Goodchild and M.G. Keane are with the Department of Gastroenterology, University College London Hospital, UK.
K. Gurusamy and B. Davidson are with the Division of Surgery and Interventional Science, University College London, UK.
S. P. Pereira is with the Institute for Liver and Digestive Health, University College London, UK.
D. C. Barratt and S. P. Pereira are joint senior authors.

routinely explain the content of 2D images to trainees. Therefore, verbally describing US images during a procedure is a relatively simple and natural task for clinicians, whereas producing labels from these descriptions eliminates the greater challenge of reviewing long video sequences and identifying anatomical landmarks *a posteriori*. In ultrasound, Jianbo et al. [24] proposed a cross-modal contrastive learning to model the correspondence between video and voice. It is shown that this self-supervised approach can be used to extract meaningful representations and improve the performance of two sub-tasks in fetal ultrasound: standard plane detection and eye-gaze prediction

To the best of our knowledge, the use of live voice commentaries for EUS identification of anatomical landmarks has not been investigated before. The aims of this work were to develop and evaluate a multi-modal neural network that learns simultaneously from US images and clinician voice commentaries given during EUS examinations, and to investigate the potential of using voice commentaries to label US images, thus reducing the burden of manual labelling. We demonstrate that real-time noisy intraoperative voice commentaries can provide an easy way to obtain US labels, even when using a small dataset.

## II. METHODS

### A. Network Architecture

To solve the labelling task, we propose to use a multi-modal approach to classify US images. Our network consists of two independent branches (one for image and one for voice) and a joint function (see Fig. 1). Each branch comprises one CNN with randomly initialised weights. The image branch takes a stack of consecutive US images as input, and the voice branch takes a voice signal converted to a spectrogram. The two branches are merged using the dot product. In all convolutions, we use a kernel size of 3 and a stride of 2 during max pooling. The problem is formalised as follows. Let $V$ denote the voice input space, $U$ the US image input space and $S$ the set of possible image labels, we train the network with a training set $T = \{(v_n, u_n, s_n)\}_{n=1}^{N}$, where $v_n \in V$, $u_n \in U$, $s_n \in S$ and $N$ the number of training samples. The model learns the mapping $f : (V, U) \rightarrow S$. In the following sections, details of the two branches and the joint function are given.

#### 1) Voice Branch

The voice branch transforms the features from the voice input space $V$ into a latent space $P$. The input of this branch is a spectrogram generated from a 1 second audio sample taken at the time a landmark was mentioned by the clinician and synchronised with the image branch. Details on the data acquisition are given in Section III.A. This branch comprises six 1D convolutions in total, each followed by a batch normalisation and a ReLu activation.

#### 2) Image Branch

The image branch transforms the features from the image input space $U$ into a latent space $Q$. This branch is based on the VGG16 network [25], a well-established architecture widely used for image-based processing in multi-branch models [19],

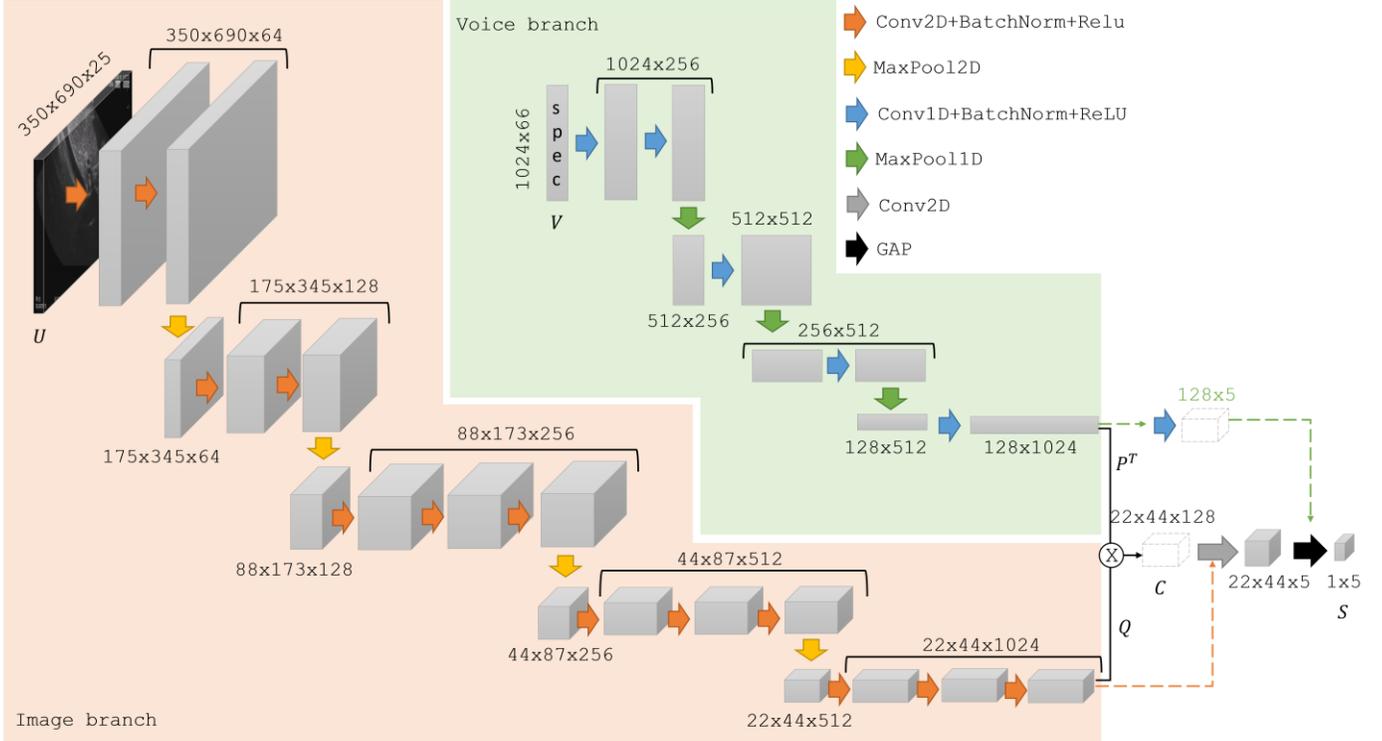

Fig. 1. Overview of the proposed network architecture, comprising two branches: image branch (orange area) and voice branch (green area). The voice branch ($P^T$) is joined with the image branch ($Q$) using the matrix product as a joint function (dot product) to obtain $C$. The image branch employs 2D convolutions (Conv2D) followed by a batch normalization (BatchNorm) and a ReLu activation (Relu), with 2D max pooling (MaxPool2D). The voice branch employs 1D convolutions (conv1D) followed by a batch normalization and a ReLu activation, with 1D max pooling (MaxPool1D). Global average pooling is denoted as GAP. The output of the network is a vector with the probabilities for each class ($S$). Ablation studies in which only one branch is used are shown with dotted arrows.

[21]. The input of this branch is a collection of 25 US images, which corresponds to 1 second of imaging data, synchronised in time with the input of the voice branch (details described in Section III.A). The image branch model comprises 13 2D convolutions in total, but unlike VGG16, each convolution is followed by a batch normalization [26] and with no fully connected layer to keep the spatiotemporal coordinates of the feature map before joining with the voice branch.

### 3) Joint Function and Decoder

The voice and image latent features $P$ and $Q$ are combined to predict the class labels $S$. The matrix product of $Q$ and $P^T$ is used as a joint function, as it is a simple parameter-free operator that allows to capture the multivariate correlation between the two high-dimensional features, and is defined as:

$$C = QP^T$$
$$C_{i,j,k} = \sum_{r=1}^{R} Q_{i,j,r} \times P^T_{r,k}$$
(1)

Where $C$ is the combined map, $i, j, k$ and $r$ denote the spatiotemporal coordinates of the feature maps, and R is the number of channels (R=1024). $C$ represents the spatiotemporal relationship between the image and voice branches, and is followed by a 2D convolution and a global average pooling (GAP) to obtain the activations for each label. GAP was chosen instead of a fully-connected layer to reduce the number of parameters and avoid overfitting. As loss function, we use a standard weighted categorical cross-entropy loss on the softmax activations. The loss function is defined as:

$$\mathcal{L} = -\frac{1}{N} \sum_{k=1}^{K} \sum_{n=1}^{N} w_k \times s_n^k \times \log(f_\theta(v_n, u_n)_k),$$
(2)

Where $N$ is the number of training samples, $K$ the number of labels (classes), $w_k$ the weight for class $k$, $s_n^k$ the target label for the training example $n$ for class $k$ and $f_\theta$ is the trained model with weights $\theta$.

### B. Network Evaluation and Metrics

To evaluate the performance of the network, a leave-one-patient-out cross-validation strategy was employed where one patient was omitted from the training dataset and the network was trained and then tested on the omitted patient. A single vector combining the predicted labels for each fold is used to evaluate the performance of the models.

We calculate the true positives ($TP_i$), the false positives ($FP_i$) and false negatives ($FN_i$) for each class, where $i$ indicates the class. We report the subset accuracy $\left(\frac{\sum_{i=1}^{n} TP_i}{m}\right)$, which corresponds to the total number of correct predictions among all predictions, the averaged precision ($\frac{1}{n}\sum_{i=1}^{n} \frac{TP_i}{(TP_i+FP_i)}$), recall ($\frac{1}{n}\sum_{i=1}^{n} \frac{TP_i}{(TP_i+FN_i)}$) and F1-score ($\frac{1}{n}\sum_{i=1}^{n} \frac{2*recall_i*precision_i}{recall_i+precision_i}$), where $m$ is the number of samples and $n$ the number of labels (in our study, $m = 143$ and $n = 5$). In addition to these metrics, we also report the confusion matrix normalised by the ground truth class.

To further investigate the accuracy uncertainty due to intra-observer variability, we performed a bootstrap sampling analysis, with 100 subsets of 10 patients each, and we report the mean and standard deviation of the accuracies obtained. The accuracy can be used as an indication of how many images would need to be manually corrected after the labelling. Note that the clinical impact of this work compared to a manual approach would need further investigation in a future large-scale study, and may vary between applications.

## III. EXPERIMENTS

### A. Data

EUS images and verbal comments from a clinician were acquired from 12 patients who underwent an EUS-guided examination to identify abnormalities in the pancreas at University College London Hospital (UCLH). Each procedure took between 11 and 29 minutes. The clinician was asked to comment on anatomical landmarks visible in the EUS field of view during the examination, following standards used when training a junior clinician and without disrupting routine clinical workflow. Anatomical landmarks varied considerably between patients and multiple landmarks were often present in the field of view. Where multiple landmarks were visible, the clinician mentioned only one – typically the landmark most relevant to the examination. We selected the 5 anatomical landmarks that had been identified in at least 7 different exams, namely: the pancreas, the portal vein, the pancreatic duct, the portal venous confluence (PVC), and the bile duct. A summary of the number of available labels for each patient and the study totals are shown in Table I. Each sample is a pair of 1 second of voice data and 25 consecutive US frames. Acquisition details are described in the next subsections.

### 1) Acquisition of US Images

US data were acquired from a Hitachi Preirus EUS console and a Pentax EG-3270UK (slim) or EG-3870UTK standard US video endoscopes, both linear and with a 7.5 MHz probe. Frames were recorded with a resolution of 720×480 pixels at an acquisition rate of 25 frames per second with a Elgato Video Capture card (Corsair GmbH, Germany). The images were cropped to 690×350 pixels for normalisation and to remove unnecessary background information such as time, data, gain, etc. The pixel size ranged from 0.09 to 0.21 mm on the $x$-axis and from 0.21 to 0.19 mm on the $y$-axis.

### 2) Acquisition of Voice Commentaries

At the same time as the US images were acquired, voice commentaries of a single gastroenterologist and/or registrar were recorded using an EVIDA digital voice recorder (EVIDA, China) with a sample rate of 48,000 Hz. Across all 12 patients, a total of 4 different clinicians were recorded (ranging from experienced consultant gastroenterologists to specialist registrars undergoing training).

### 3) Voice and US Data Synchronization and Processing

Imaging data and voice data were synchronised using a reference timestamp. The audio samples where a clinician mentioned a label were manually identified and collated with the corresponding synchronised images. A total of 143 pairs of image and voice sets were identified, as summarised in Table I. An illustration of the voice sample durations is shown in Fig. 2.

For normalisation, each voice sample was cropped or padded with zeros such that the final duration was 1 second and each set of images contained 25 frames. Each voice sample was converted to a spectrogram of size 1024×66 by computing a 2046-point Short-time Fourier Transform (STFT) with a window length of 1200 frames (25 milliseconds) and a step of 720 frames (40% overlap).

TABLE I
SUMMARY OF THE DATA ACQUIRED FROM THE 12 PATIENTS. THE TABLE SHOWS THE NUMBER OF SAMPLES FOR EACH OF THE 5 ANATOMICAL LANDMARKS (PANCREAS, PORTAL VEIN, PANCREATIC DUCT (PD), PORTAL VENOUS CONFLUENCE (PVC) AND BILE DUCT) AND EACH PATIENT, INCLUDING A SUMMARY OF THE TOTAL NUMBER PER PATIENT AND LABEL. EACH SAMPLE IS A PAIR OF 1 SECOND OF VOICE DATA AND 25 US FRAMES

| Label | Samples | Patient 1 | 2 | 3 | 4 | 5 | 6 | 7 | 8 | 9 | 10 | 11 | 12 |
|---|---|---|---|---|---|---|---|---|---|---|---|---|---|
| Pancreas | 16 | 0 | 0 | 0 | 1 | 1 | 0 | 0 | 5 | 4 | 2 | 2 | 1 |
| Portal vein | 24 | 0 | 1 | 1 | 1 | 1 | 0 | 0 | 8 | 5 | 0 | 6 | 1 |
| PD | 38 | 1 | 11 | 0 | 2 | 5 | 4 | 0 | 5 | 2 | 3 | 2 | 3 |
| PVC | 24 | 0 | 2 | 1 | 1 | 3 | 1 | 0 | 7 | 8 | 0 | 1 | 0 |
| Bile duct | 41 | 1 | 4 | 2 | 3 | 8 | 2 | 1 | 2 | 10 | 3 | 5 | 0 |
| Total samples | 143 | 2 | 18 | 4 | 8 | 18 | 7 | 1 | 27 | 29 | 8 | 16 | 5 |

### B. Implementation Details

The proposed model was implemented using Tensorflow 2.4. We trained the model on 200 epochs with a batch size of 8 and an Adam optimiser with a learning rate of 0.001 with the following class weights: pancreas = 2.6, portal vein = 1.7, pancreatic duct = 1.1, PVC = 1.7 and bile duct = 1.0. Each patient-fold was run on a GPU GeForce GTX 1080 for approximately 12 hours.

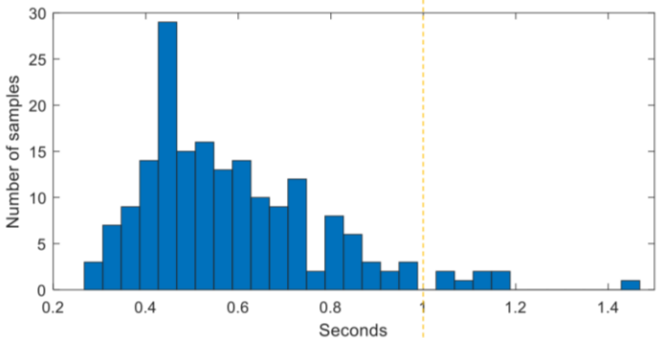

Fig. 2. Histogram depicting the length of all voice samples (in seconds). The vertical dashed line indicates the 1 second threshold to which all network inputs were cropped or padded (for normalization purposes).

### C. Study Objectives

Due to the limited number of patients, experiments were carefully designed to avoid overfitting and information bleeding [27], while maintaining a large enough dataset to evaluate the label prediction task using voice. Therefore, extensive fine-tuning of the proposed network was not adopted, and performance was evaluated using a leave-one-patient-out cross validation strategy. Rather than achieving optimum performance, which would be unfeasible with the current dataset, our experiments were intended to investigate the following hypotheses and research questions:

#### 1) Pre-trained weights

This experiment aims to test if pre-trained weights for both voice and image improve the performance of the network. The adapted network architecture used in this experiment is shown in Fig. 3.

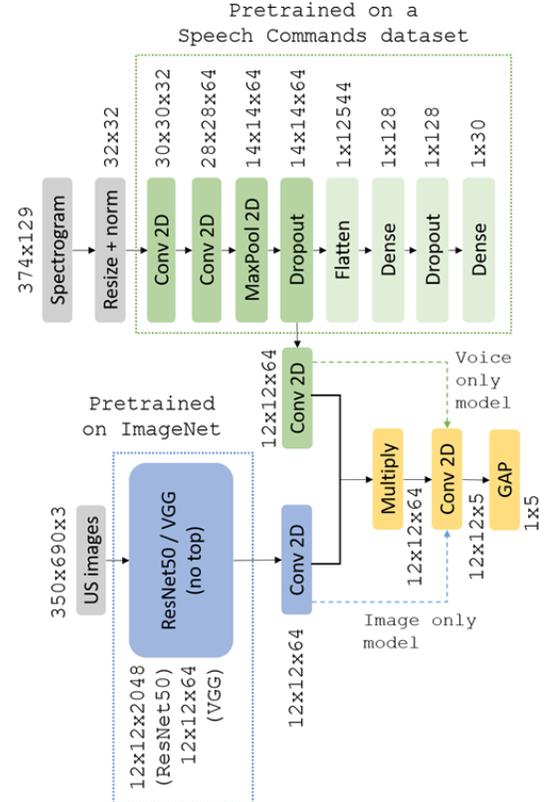

Fig. 3. Network architecture using pre-trained weights for both image (blue) and voice (green) branches. Dotted arrows denote ablation studies when only one of the branches was used. Dotted rectangles denote layers with pre-trained weights.

For the voice branch, we pre-trained a simple audio recognition model on a public dataset for singe-word command speech recognition [28]. This dataset is composed of 64,727 one-second raw audio files, each containing a single command from a set of 30 commands (e.g., 'go', 'left'). Each voice sample was converted to a spectrogram of size 374×129 by computing a STFT with a window length of 255 frames and a step of 128 frames. To train the model faster, the spectrogram is first resized (to 32x32 pixels) and then the input is normalised with a zero mean and standard deviation of one. We froze all pre-trained layers during training and we added a 2D convolution after the Dropout layer in order to join the voice branch with the image branch.

For the image branch, we used a VGG16 [25] and a ResNet50 [29], both trained on ImageNet [30]. We removed the top of the pre-trained network and we froze all layers. To join the image branch with the voice branch, we added a 2D convolutional network.

To join the two branches, we used the same join function as the proposed model and added a 2D convolutional layer and a GAP to enable the comparison with the proposed model. We trained these models (VGG16 and ResNet50) on 200 epochs with a batch size of 8 and an Adam optimiser with a learning rate of 1e$^{-5}$ and we used the same class weights and loss function.

*2) Voice-only and Image-only Models*

In order to understand the contribution of each branch, a feature ablation study was designed to quantitatively evaluate the importance of the voice and image branches when used alone (i.e. the labelling performance with no image or voice input). This is a challenging task to perform with the image-only branch and our dataset, because images could contain multiple landmarks, but are only assigned the single most relevant label, as the clinician can only mention one landmark at a time. Therefore, the performance of this branch alone was expected to be low. We tested this approach using both the proposed architecture and the architecture with pre-trained weights. For both approaches, we kept the architecture fixed by skipping the multiplication between the image and voice branches as represented in Fig. 1 and Fig. 3. Table II shows the number of epochs and learning rate used for training the different models.

TABLE II
NUMBER OF EPOCHS AND LEARNING RATE (LR) IN THE ABLATION STUDY

| Model | Epochs | LR |
|---|---|---|
| Voice-only not pretrained | 200 | 1e$^{-6}$ |
| Voice-only with pretrained weights | 1000 | 1e$^{-3}$ |
| Image-only not pretrained | 80 | 1e$^{-5}$ |
| Image-only with VGG16 with pret. weights | 200 | 1e$^{-5}$ |
| Image only with ResNet50 with pret. weights | 200 | 1e$^{-5}$ |

*3) Random Pairs of Image and Voice*

In this experiment, we investigate the effect of training the model using randomised matched pairs. In each epoch, we train the model with a training set $T = \{(v_i, u_j, s_n)\}_{n=1}^{N}$ such that the class $s_i = s_j = s_n$, where $v_i \in V$, $u_j \in U$, $s_n \in S$. $N$ is the number of training samples.

*4) Reduced Image Input*

In order to assess the impact due to potentially smaller input image size, we evaluated the model using 3 US frames as input (equivalent to 3 channels, as it is a widely used configuration in most VGG and ResNet50 architectures), randomly selected at each epoch from the 25 frames available for each sample.

### D. Class Activation Maps

We are also interested in the visual explanation and spatial localisation of important regions in the US images that were used to predict the class.

We used the Gradient-weighted Class Activation Mapping (Grad-CAM) to generate the class activation maps for each sample [31]. These maps provide an insight into the model interpretation by backpropagating the gradients from the last convolutional layer. In this work, we explore the importance of these techniques in clinical practice, using the proposed voice labelling models. In particular, we show that these model interpretations may be useful for safeguarding against potential misclassification *in vivo*, provide insight into regions that have been used for correct predictions, and suggest previously overlooked landmarks for guidance purposes. The latter two are valuable sources of information that can be incorporated into model training to combat practical limitations such as data availability. Perhaps more importantly, the class activation map analysis may also establish confidence that the proposed deep-learning models have learned relevant, application-specific features; as opposed to arbitrary, unstructured regions commonly associated with overfitting in models.

## IV. RESULTS AND DISCUSSION

The quantitative results obtained with the different models in terms of accuracy, precision, recall and F1-score are shown in Table III. The best performance was obtained with the model trained on both image and voice without using pre-trained weights. Although the models with pre-trained weights performed better in the ablation studies in which only one branch was used, this was not the case when both branches were used. We attribute this to the fact that the spatiotemporal correlation between image and voice is lost when training both branches independently.

One of the challenges of our dataset is that several landmarks are situated near to each other and often appear simultaneously in the same image. The examining clinician may only mention one landmark of interest, or alternatively report multiple landmarks in succession while maintaining the same view. This was a common occurrence for pancreas, pancreatic duct and PVC images as shown in Fig. 4. The pancreatic duct is located within the pancreatic parenchyma and will always appear in images where the pancreatic head, body or tail are also visible. The pancreatic head and common bile duct were also susceptible, because the pancreatic head, duct and the ampulla are all in close proximity. This is supported by the lower accuracy obtained with the image-only ablation study, where the image branch was considered independently (accuracy of 0.20 with no pre-trained weights and accuracy of 0.31 with ResNet50 and pre-trained weights), as most of the images contained multiple landmarks. Further investigation with an experienced clinician showed that some of the misclassified images were classified as one of the other landmarks seen in the image. For instance, in Fig. 4 we can see examples of images labelled as PVC but classified as pancreas or pancreatic duct, which are also visible on the images.

TABLE III
QUANTITATIVE RESULTS FOR THE DIFFERENT MODELS TESTED IN TERMS OF ACCURACY, PRECISION, RECALL AND F1-SCORE.

| Model | Accuracy | Precision | Recall | F1-score |
|---|---|---|---|---|
| Proposed with image + voice (no pretrained weights) | **0.76** | **0.74** | **0.74** | **0.74** |
| Proposed with VGG16 + pretrained weights | 0.66 | 0.60 | 0.59 | 0.59 |
| Proposed with ResNet50 + pretrained weights | 0.59 | 0.54 | 0.54 | 0.54 |
| Image branch only (not pretrained) | 0.20 | 0.20 | 0.34 | 0.24 |
| Image branch only with VGG16 + pret. weights | 0.28 | 0.24 | 0.24 | 0.23 |
| Image branch only with ResNet50 + pret. weights | 0.31 | 0.26 | 0.26 | 0.26 |
| Voice branch only (not pretrained) | 0.42 | 0.40 | 0.40 | 0.38 |
| Voice branch only pretrained | 0.66 | 0.68 | 0.65 | 0.65 |
| Proposed with random pairs | 0.70 | 0.67 | 0.67 | 0.67 |
| Proposed with reduced image input | 0.66 | 0.63 | 0.63 | 0.62 |

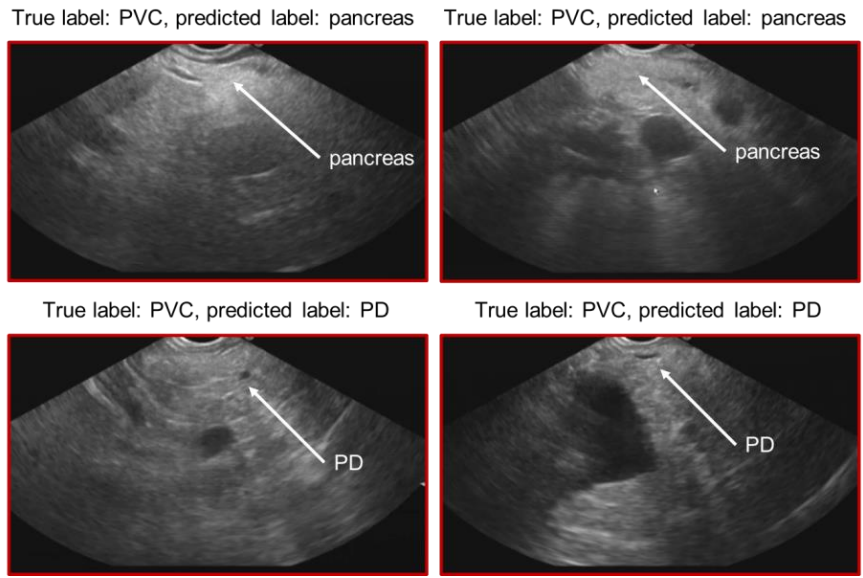

Fig. 4. Visual examples of cases of portal venous confluence (PVC) which have been wrongly classified as pancreas or pancreatic duct (PD), which are also visible on the images.

Considering a voice-only model, the accuracy was also lower (accuracy of 0.66 with pre-trained weights) compared to using both voice and image branches. The poor performance of this branch can be associated with the use of intraoperative voice recordings, as labels can be described differently due to the lack of a standard protocol when mentioning a label (e.g., pancreatic duct can be described as 'pancreatic duct' or 'PD'). This could be further explored by including a protocol when a label is mentioned in the clinical workflow, however, one of the aims of this research is to integrate the proposed method in a natural manner, which includes no alteration of the current workflow. The quantitative results obtained in the ablation study with the image-only and voice-only models confirm the importance of a multi-modal approach.

The bootstrapping analysis on the proposed model, showed an accuracy of 0.76 with a standard deviation of 0.04.

Results using random pairs of image and voice showed a slight decrease of all metrics (accuracy of 0.76 for paired image and voice and accuracy of 0.70 for random pairs of image and voice). This finding suggests that training the model with random pairs of image and voice may add data redundancy into the network.

To investigate the effect of the number of images as input, as a potential strategy to reduce the size of the inputs, we evaluated the model with 3 ultrasound images as input. Results show that a reduced number of images may adversely affect the accuracy (0.76 versus 0.66, for the proposed model, and the model with reduced input, respectively).

In Fig. 5 we show the confusion matrix obtained with the proposed model. We can observe that the PVC is the structure with a lower accuracy, as the model classified 17% of the images as pancreatic duct, and 12% of the images as bile duct (see Fig. 4 for visual examples).

Fig. 5. Confusion matrix obtained with the proposed method normalised by true label. The *x*-axis corresponds to the predicted labels and the *y*-axis to the ground truth label. Dark blue represents higher accuracy. The portal venous confluence is denoted as PVC.

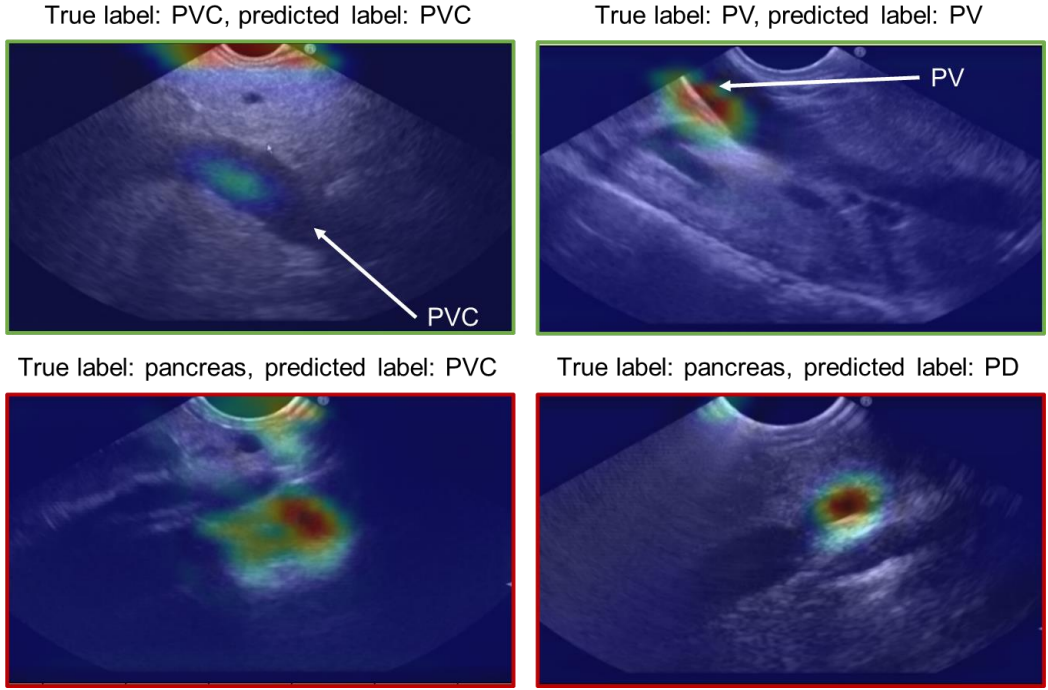

Fig. 6. Visual examples of class activation maps. The top row shows the map of correctly classified cases of portal venous confluence (PVC) and portal vein (PV). The bottom row shows the map of incorrectly classified pancreas as PVC on the left, an pancreatic duct (PD) on the right.

Further qualitative analysis on the class activation maps showed that the model is capable of learning features from the US images focusing on anatomical structures. For instance, in Fig. 6 (top left) we can see that the area in the middle of the image, where the PVC is shown, activated the positive prediction of PVC, and in Fig. 6 (top right) the region defining the edge of portal vein activated a positive prediction of portal vein. These qualitative observations demonstrate anatomically meaningful feature representations learned by the networks as regions of interest directly relevant to the classification were extracted. This feature relevance is evidence that overfitting did not occur during training, despite the small data set used in our study. We also generated class activation maps when the wrong pair of image label and voice label was given. We observed that in some cases, the model was able to focus on different regions of the image depending of the voice label that was given. As an example, Fig. 7 (top) shows the map of an image that corresponds to the PVC when is paired with a voice label of bile duct (left) or portal vein (right). We can observe that the map focuses on similar regions but for bile duct is more intense that for portal vein. Fig. 7 (bottom) shows an example of an image corresponding to the pancreatic duct paired with voice labels of PVC and bile duct. While the voice label of PVC does not seem to use any region of the image to make the prediction, when the bile duct voice label is given, it uses the area of the pancreatic duct to make the prediction of bile duct. The two anatomical landmarks may look similar on the US images as both are ducts.

The main limitation of this study was the small amount of labelled time-series data available. To the best of our knowledge, no public US datasets with voice commentaries exist. With the current dataset, a division of the patient cases into a training, validation and testing sets is challenging, as the distribution of the labels among the patients is varied (i.e., not all patients have all the labels and the number of labels is different for each set) as shown in Table I. Therefore, we decided against using a hold-out dataset for validation.

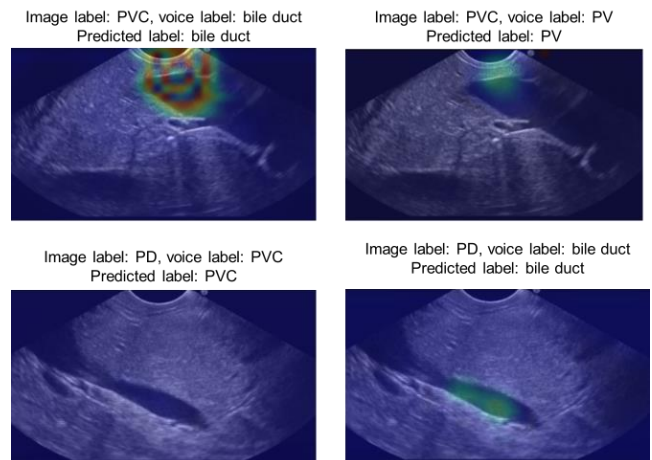

Fig. 7. Visual examples of class activation maps when the wrong pair of image label and voice label is given. The portal venous confluence is denoted as PVC, portal vein as PV and pancreatic duct as PD.

Deep learning-based methods have been used previously to classify US images. For instance, Cheng et al. [5], used transfer learning to classify abdominal US images into 11 categories with high accuracy (77.9%) using a VGGNet approach. In this study, the ground truth classification was obtained from text annotations and reviewed by an experienced radiologist. Results were compared to a human classifier, in which a different radiologist spent approximately 12 hours retrospectively labelling a test set with 1,423 images using a custom graphical user interface. The accuracy obtained by the human classifier was 71.7%. In our study, we label 3,575 images (143 sets of 25 frames each) with an accuracy of 76%. A direct comparison of the results is not possible, as we studied a different dataset and a different application. Even so, our results demonstrate the principle of using real-time voice commentary during image acquisition as an addition to the images themselves for EUS classification to substantially reduce labelling time and increase the accuracy and speed of structure by all levels of endosonographer. Using our approach, key images containing anatomical structures which are of interest to the clinician can be efficiently labelled (at the image level) and used to train classification models. Such a model would enhance classification of key structures so that pathological tissues can be detected rapidly and biopsied if necessary. If biopsy results were unexpected, labelled non-pathological structures could be efficiently located and reviewed in subsequent EUS scans. A clinical study to evaluate our approach is proposed for future work.

## V. Conclusion

This study proposes a multi-modal learning method to automatically label EUS images during procedures in which manual labelling is considered a difficult task. Voice comments are given in real-time by a clinician during EUS procedures. Results show an improved efficiency when using voice over the use of images only. The impact of this work is expected to be considerable in both clinical and research environments in the long term, especially given difficulties in obtaining labelled US data for deep learning methods, currently considered a major bottleneck. Future work will focus on data gathering and in the next steps to fully automate the method proposed, for example, by using a speech recognition system to filter the labels of interest.

Code will be available at: https://ebonmati.github.io/.